\documentclass{llncs}

\usepackage{bbding} 

\usepackage{cite}
\usepackage{amsmath,amsfonts}
\usepackage{bbm}
\usepackage{amssymb}
\usepackage{mathrsfs}
\usepackage{algorithmic}
\usepackage[ruled, lined, longend, linesnumbered]{algorithm2e}
\usepackage{textcomp}
\usepackage{xcolor}
\usepackage{tikz}
\usepackage{makecell}
\usepackage{pgfplots}
\usepackage{xurl}
\usepackage{epstopdf}
\usepackage{moresize}
\usepackage{listings}
\usepackage{color}
\usepackage{transparent}
\usepackage{multirow}
\usepackage{pgfplots}
\usepackage{comment}

\newcommand{\ignore}[1]{}

\definecolor{lbcolor}{rgb}{0.9,0.9,0.9}
\definecolor{darkblue}{rgb}{0, 0.125, 0.576}
\definecolor{gray}{rgb}{0.5,0.5,0.5}
\definecolor{darkgreen}{rgb}{0,0.5,0}
\definecolor{darkred}{rgb}{0.6,0,0}
\definecolor{mygreen}{rgb}{0,0.6,0}
\definecolor{mygray}{rgb}{0.5,0.5,0.5}
\definecolor{mymauve}{rgb}{0.58,0,0.82}
\definecolor{dkgreen}{rgb}{0,0.6,0}

\definecolor{lightgray}{rgb}{0.85,0.85,0.85}
\definecolor{lightgreen}{rgb}{0.7,0.9,0.7}
\definecolor{lightblue}{rgb}{0.7,0.7,0.9}
\definecolor{lightred}{rgb}{0.9,0.7,0.7}
\definecolor{OrangeRed}{rgb}{1.0, 0.27, 0.0}

\lstdefinelanguage{diff}{
	frame=single,  
	basicstyle=\ttfamily\footnotesize,
	morecomment=[f][\color{darkgreen}]{@@},
	morecomment=[f][\color{darkblue}]{+\ },
	morecomment=[f][\color{OrangeRed}]{-\ },
	keywordstyle=\color{black},
	morekeywords={static, void, if, else, while, struct, int, true, false, unsigned, long, return, goto},   
	framexleftmargin=0.5em,
}

\usepackage{pifont}

\newcommand{\DBIAS}{\mathsf{dBias}}

\begin{document}

\title{Certifying the Fairness of KNN in the Presence of Dataset Bias\thanks{
This work was partially funded by the U.S.\ National Science Foundation grants CNS-1702824, CNS-1813117 and CCF-2220345.}}

\author{Yannan Li\ \Envelope
   \and Jingbo Wang
   \and Chao Wang
}

\authorrunning{}

\institute{University of Southern California, Los Angeles CA 90089, USA\\
\email{\{yannanli,jingbow,wang626\}@usc.edu}
}

\maketitle

\begin{abstract}
We propose a method for certifying the fairness of the classification result of a widely used supervised learning algorithm, the $k$-nearest neighbors (KNN), under the assumption that the training data may have historical bias caused by systematic mislabeling of samples from a protected minority group. 
To the best of our knowledge, this is the first certification method for KNN based on three variants of the fairness definition: individual fairness, $\epsilon$-fairness, and label-flipping fairness. 
We first define the fairness certification problem for KNN and then 
propose sound approximations of the complex arithmetic computations used in the state-of-the-art KNN algorithm.  
This is meant to lift the computation results from the concrete domain to an abstract domain, to reduce the computational cost. 
We show effectiveness of this \emph{abstract interpretation} based technique through experimental evaluation on six datasets widely used in the fairness research literature.  We also show that the method is accurate enough to obtain fairness certifications for a large number of test inputs, despite the presence of historical bias in the datasets. 
\end{abstract}

\section{Introduction}
\label{sec:intro}

Certifying the fairness of the classification output of a machine learning model has become an important problem.  This is in part due to a growing interest in using machine learning techniques to make socially sensitive decisions in areas such as education, healthcare, finance, and criminal justice systems. 
One reason why the classification output may be biased against an individual from a protected minority group  is because the dataset used to train the model may have historical bias; that is, there is systematic mislabeling of samples from the protected minority group.  
Thus, we must be extremely careful while considering the possibility of using the classification output of a machine learning model, to avoid perpetuating or even amplifying historical bias.

One solution to this problem is to have the ability to certify, with certainty, that the classification output $y=M(x)$ for an individual input $x$ is fair, despite that the model $M$ is learned from a dataset $T$ with historical bias. 
This is a form of \emph{individual fairness} that has been studied in the fairness literature~\cite{DworkHPRZ12};  it requires that the classification output remains the same for input $x$ even if historical bias were not in the training dataset $T$.
However, this is a challenging problem and, to the best of our knowledge, techniques for solving it efficiently are still severely lacking.  Our work aims to fill the gap.

Specifically, we are concerned with three variants of the fairness definition.  
Let the input $x = \langle x_1,\ldots,x_D \rangle$ be a $D$-dimensional input vector, and $\mathcal{P}$ be the subset of vector indices corresponding to the \emph{protected} attributes (e.g., race, gender, etc.).  
The first variant of the fairness definition is \emph{individual fairness}, which requires that similar individuals are treated similarly by the machine learning model.  For example, if two individual inputs $x$ and $x'$ differ only in some protected attribute $x_i$, where $i\in\mathcal{P}$, but agree on all the other attributes,  the classification output must be the same. 
The second variant is \emph{$\epsilon$-fairness}, which extends the notion of individual fairness to include inputs whose un-protected attributes differ and yet the difference is bounded by a small constant ($\epsilon$).  
In other words, if two individual inputs are almost the same in all unprotected attributes, they should also have the same classification output. 
The third variant is \emph{label-flipping fairness}, which requires the aforementioned fairness requirements to be satisfied  even if a biased dataset $T$ has been used to train the model in the first place.  
That is, as long as the number of mislabeled elements in $T$ is bounded by $n$, the classification output must be the same.

We want to certify the fairness of the classification output for a popular supervised learning technique called the $k$-nearest neighbors (KNN) algorithm.  
Our interest in KNN comes from the fact that, unlike many other machine learning techniques, KNN is a \emph{model-less} technique and thus does not have the high cost associated with training the model.  Because of this reason, KNN has been widely adopted in real-world applications~\cite{guo2003knn,andersson2020predicting,wu2011drex,adeniyi2016automated,narudin2016evaluation,firdausi2010analysis, li2007network,su2011real,xie2012scalable}. 
However, obtaining a fairness certification for KNN is still challenging and, in practice, the most straightforward approach of \emph{enumerating all possible scenarios} and then \emph{checking if the classification outputs obtained in these scenarios agree} would have been prohibitively expensive.

\begin{figure}[t]
\centering
\includegraphics[width=0.75\textwidth]{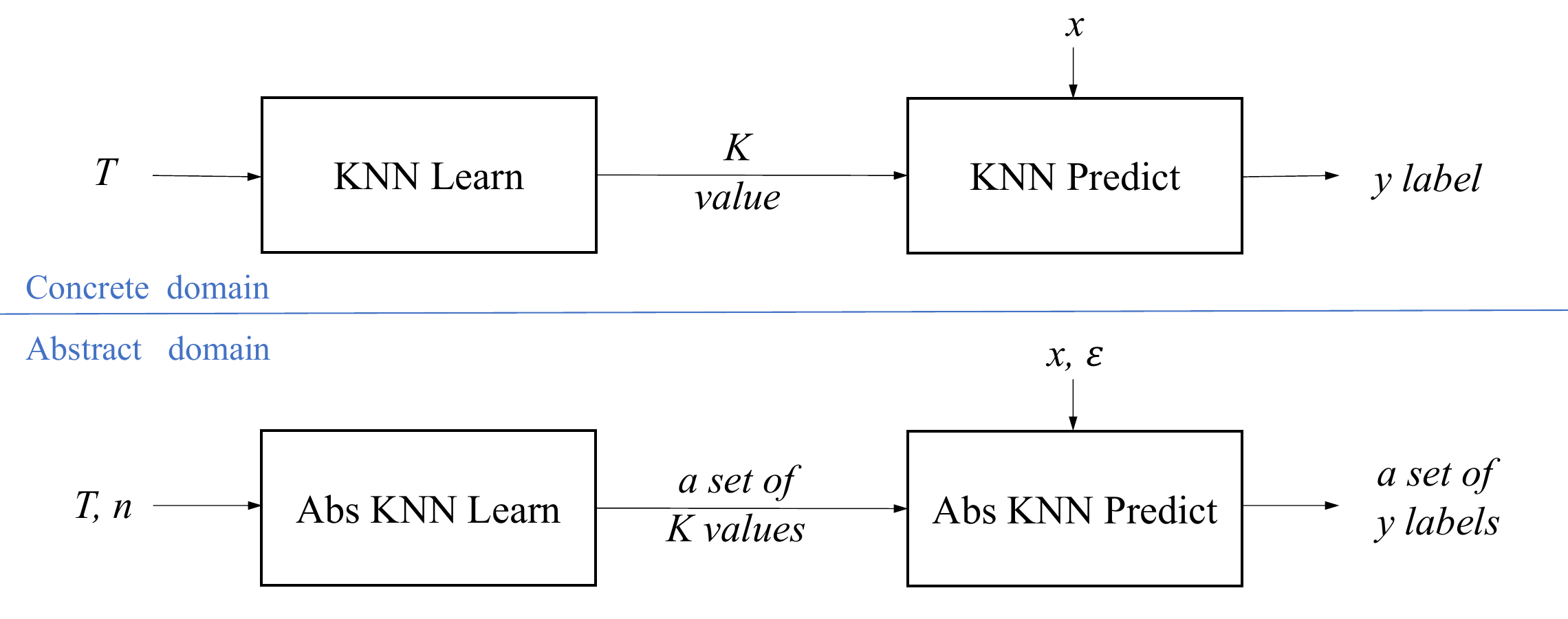}
\caption{\textsc{FairKNN}: our method for certifying fairness of KNNs with label bias.}
\label{fig:flow}
\end{figure}

To overcome the challenge, we propose an efficient method based on the idea of \emph{abstract interpretation}~\cite{CousotC77}.  Our method relies on sound approximations to analyze the arithmetic computations used by the state-of-the-art KNN algorithm both accurately and efficiently.
Figure~\ref{fig:flow} shows an overview of our method in the lower half of this figure, which conducts the analysis in an abstract domain, and the default KNN algorithm in the upper half, which operates in the concrete domain.
The main difference is that, by staying in the abstract domain, our method is able to analyze a large set of possible training datasets (derived from $T$ due to $n$ label-flips) and a potentially-infinite set of inputs (derived  from $x$ due to $\epsilon$ perturbation) symbolically,  as opposed to analyze a single training dataset and a single input concretely.

To the best of our knowledge, this is the first method for KNN fairness certification in the presence of dataset bias. 
While Meyer et al.~\cite{MeyerAD21,abs-2206-03575} and Drews et al.~\cite{Drews2020PLDI} have investigated robustness certification techniques, their methods target decision trees and linear regression, which are different types of machine learning models from KNN. 
Our method also differs from the KNN data-poisoning robustness verification techniques developed by Jia et al.~\cite{jia2022certified} and Li et al.~\cite{li2022proving}, which do not focus on \emph{fairness} at all; for example, they do not distinguish \emph{protected} attributes from \emph{unprotected} attributes.
Furthermore, Jia et al.~\cite{jia2022certified} consider the prediction step only while ignoring the learning step, and 
 Li et al.~\cite{li2022proving} do not consider  label flipping.  
Our method, in contrast, considers all of these cases.

We have implemented our method and demonstrated the effectiveness through experimental evaluation. We used all of the six popular datasets in the fairness research literature as benchmarks. 
Our evaluation results show that 
the proposed method is efficient in analyzing complex arithmetic computations used in the state-of-the-art KNN algorithm, and is accurate enough to obtain fairness certifications for a large number of test inputs. 
To better understand the impact of historical bias, we also compared the fairness certification success rates across different demographic groups. 

To summarize, this paper makes the following contributions:
\begin{itemize}
\item 
We propose an abstract interpretation based method for efficiently certifying the fairness of KNN classification results in the presence of dataset bias.  The method relies on sound approximations to speed up the analysis of both the learning and the prediction steps of the state-of-the-art KNN algorithm, and is able to handle three variants of the fairness definition.
\item
We implement the method and evaluate it on six datasets that are widely used in the fairness literature, to demonstrate the efficiency of our approximation techniques as well as the effectiveness of our method in obtaining sound fairness certifications for a large number of test inputs. 
\end{itemize}

The remainder of this paper is organized as follows.  We first present the technical background in Section~\ref{sec:background} and then give an overview of our method in Section~\ref{sec:method-overall}.  Next, we present our detailed algorithms for certifying the KNN prediction step in Section~\ref{sec:abs-knn-prediction} and certifying the KNN learning step in Section~\ref{sec:abs-knn-learning}.  This is followed by our experimental results in Section~\ref{sec:experiment}. We review the related work in Section~\ref{sec:related} and, finally, give our conclusion in Section~\ref{sec:conclusion}.

\section{Background}
\label{sec:background}

Let $L$ be a supervised learning algorithm that  takes the training dataset $T$ as input and returns a learned model $M=L(T)$ as output.   
The training set $T = \{(x,y)\}$ is a set of labeled samples, where each $x \in \mathcal{X} \subseteq \mathbb{R}^D$ has $D$ real-valued attributes, and the $y \in \mathcal{Y} \subseteq \mathbb{N}$ is a class label. 
The learned model $M : \mathcal{X} \rightarrow \mathcal{Y}$ is a function that returns the classification output $y'\in\mathcal{Y}$ for any input $x'\in\mathcal{X}$. 

\subsection{Fairness of the Learned Model}
\label{sec:fairness-definition-1}

We are concerned with \emph{fairness} of the classification output $M(x)$ for an individual input $x$. 
Let $\mathcal{P}$ be the set of vector indices corresponding to the protected attributes in $x\in\mathcal{X}$.  We say that $x_i$ is a protected attribute (e.g., race, gender, etc.) if and only if $i\in\mathcal{P}$.

\begin{definition}[Individual Fairness]
For an input $x$, the classification output $M(x)$ is fair if, for any input $x'$ such that (1) $x_j \neq x'_j$ for some $j\in\mathcal{P}$ and (2) $x_i=x'_i$ for all $i\not\in\mathcal{P}$, we have  $M(x) = M(x')$.
\end{definition}

It means two individuals ($x$ and $x'$) differing only in some protected attribute (e.g., gender) but agreeing on all other attributes must be treated equally. 
While being intuitive and useful, this notion of fairness may be too narrow.  For example, if two individuals differ in some unprotected attributes and yet the difference is considered \emph{immaterial}, they must still be treated equally.  This can be captured by $\epsilon-$fairness.

\begin{definition}[$\epsilon$-Fairness]
For an input $x$, the classification output $M(x)$ is fair if, for any input $x'$ such that 
(1) $x_j \neq x'_j$ for some $j \in \mathcal{P}$
and (2) $|x_i - x'_i| \leq \epsilon$ for all $i \not\in \mathcal{P}$,
we have $M(x) = M(x')$.
\end{definition}

In this case, such inputs $x'$ form a set. 
Let $\Delta^\epsilon(x)$ be the set of all inputs $x'$ considered in the $\epsilon-$fairness definition.  That is, $\Delta^\epsilon(x) := \{x' ~|~ 
 x_j \neq x'_j \mbox{ for some } j \in \mathcal{P},
|x_i - x'_i| \leq \epsilon \mbox{ for all } i \not\in \mathcal{P} 
\}$. 
By requiring $M(x)=M(x')$ for all $x' \in \Delta^\epsilon(x)$, $\epsilon$-fairness guarantees that a larger set of individuals similar to $x$ are treated equally.

Individual fairness can be viewed as a special case of $\epsilon$-fairness, where $\epsilon=0$.
In contrast,  when $\epsilon>0$,  the number of elements in $\Delta^\epsilon(x)$ is often large and sometimes infinite.
Therefore, the most straightforward approach of certifying fairness by enumerating all possible elements in $\Delta^\epsilon(x)$ would  not work.  
Instead, any practical solution would have to rely on abstraction.

\subsection{Fairness in the Presence of Dataset Bias}
\label{sec:fairness-definition-2}

Due to historical bias, the training dataset $T$ may have contained samples whose output are unfairly labeled.  Let the number of such samples be bounded by $n$.  We assume that there are no additional clues available to help identify the mislabeled samples.
Without knowing which these samples are, fairness certification must consider all of the possible scenarios.  Each scenario corresponds to a de-biased dataset, $T'$, constructed by flipping back the incorrect labels in $T$.
Let $\DBIAS{}_n(T) = \{T'\}$ be the set of these possible de-biased (clean) datasets.  Ideally, we want all of them to lead to the same classification output.

\begin{definition}[Label-flipping Fairness]
For an input $x$, the classification output $M(x)$ is fair against label-flipping bias of at most $n$ elements in the dataset $T$ if, for all $T'\in \DBIAS{}_n(T)$,  we have $M'(x) = M(x)$ where $M'=L(T')$. 
\end{definition}

Label-flipping fairness differs from and yet complements individual and $\epsilon$-fairness in the following sense. 
While individual and $\epsilon$-fairness guarantee equal output for similar inputs, label-flipping fairness guarantees equal output for similar datasets. 
Both aspects of fairness are practically important.
By combining them, we are able to define the entire problem of certifying fairness in the presence of historical bias.

To understand the complexity of the fairness certification problem, we need to look at the size of the set $\DBIAS_n(T)$, similar to how we have analyzed the size of  $\Delta^\epsilon(x)$.  
While the size of $\DBIAS{}_n(T)$ is always finite, it can be astronomically large in practice.
Let $q$ is the number of unique class labels and  $m$ be the actual number of flipped elements in $T$. 
Assuming that each flipped label may take any of the other $q-1$ possible labels, 
the total number of possible \emph{clean sets} is ${|T|\choose m}\cdot(q-1)^m$ for each $m$.  
Since $m \leq n$, $|\DBIAS{}_n(T)| = \sum_{m = 1}^{n}{|T|\choose m}\cdot(q-1)^m$.
Again, the number of elements in $\DBIAS_n(T)$ is too large to enumerate, which means any practical solution would have to rely on abstraction.

\section{Overview of Our Method}
\label{sec:method-overall}

Given the tuple $\langle T, \mathcal{P}, n, \epsilon, x \rangle$, where $T$ is the training set, $\mathcal{P}$ represents the protected attributes, $n$ bounds the number of biased elements in $T$, and $\epsilon$ bounds the perturbation of $x$, 
our method checks if the KNN classification output for $x$ is fair.

\subsection{The KNN Algorithm}

Since our method relies on an \emph{abstract interpretation} of the KNN algorithm, we first explain how the KNN algorithm operates in the concrete domain (this subsection), and then lift it to the abstract domain in the next subsection.

\begin{figure}[!h]
\centering
\begin{minipage}{0.9\linewidth}
\begin{lstlisting}[numbers=left]
func KNN_predict((*$T, K, x$*)) {
	Let (*$T_x^K$*) = the (*$K$*) nearest neighbors of (*$x$*) in (*$T$*);
	Let (*$Freq(T_x^K)$*) = the most frequent label in (*$T_x^K$*);
	return (*$Freq(T_x^K)$*);
}

func KNN_learn((*$T$*)) { 
	for (each (*candidate $k$ value*)) {  // conducting p-fold cross validation
		Let (*$\{G_i\}$*) = a partition of (*$T$*) into (*$p$*) groups of roughly equal size;
		Let (*$err_i^k$*) = (*$\{ (x,y)\in G_i ~|~ y \neq \mbox{KNN\_predict}(T\setminus G_i, k, x) \}$*) for each (*$G_i$*);
	}
	Let (*$K$*) =  (*$\underset{k}{\arg\min}$*) (*$\frac{1}{p}\sum_{i = 1}^{p} \frac{ |err_i^k| }{ |G_i| }$*);
	return (*$K$*);
}
\end{lstlisting}
\end{minipage}
\caption{The KNN algorithm, consisting of the prediction and learning steps.}
\label{fig:KNN-algo}
\end{figure}

As shown in Fig.~\ref{fig:KNN-algo}, KNN has a prediction step where \texttt{KNN\_predict} computes the output label for an input $x$ using $T$ and a given parameter $K$, 
and a learning step where \texttt{KNN\_learn} computes the $K$ value from the training set $T$. 

Unlike many other machine learning techniques, KNN does not have an explicit model $M$; instead, $M$ can be regarded as the combination of $T$ and $K$.

Inside \texttt{KNN\_predict}, the set $T_x^K$ represents the $K$-nearest neighbors of $x$ in the dataset $T$, where distance is measured by Euclidean (or Manhattan) distance in the input vector space. 
$Freq(T_x^K)$ is  the most frequent label in $T_x^K$. 
%

Inside \texttt{KNN\_learn}, a technique called \emph{$p$-fold cross validation} is used to select the optimal value for $K$, e.g., from a set of candidate $k$ values in  the range $[1, |T|\times (p-1)/p]$ by minimizing  classification error, as shown in Line~12. 
This is accomplished by first partitioning  $T$ into $p$ groups of roughly equal size (Line~9), and then computing $err_i^k$ (a set of misclassified samples from $G_i$) by treating $G_i$ as the evaluation set, and $T\setminus G_i$ as the training set. 
Here, an input $(x,y)\in G_i$ is ``misclassified'' if the expected output label, $y$, differs from the output of \texttt{KNN\_predict} using the candidate $k$ value.


\subsection{Certifying the KNN Algorithm}

Algorithm~\ref{alg:overview} shows the top-level procedure of our fairness certification method, which first executes the KNN algorithm in the concrete domain (Lines~1-2), to obtain the default $K$ and $y$, and then starts our analysis in the abstract domain.

\begin{algorithm}[!h]
\caption{Our method for certifying fairness of KNN for input $x$.
}
\label{alg:overview}
{\footnotesize
$K$ = \texttt{KNN\_learn}($T$);\\
$y$ = \texttt{KNN\_predict}($T, K, x$);\\

$KSet$ = \textit{\texttt{abs\_KNN\_learn}}($T, n$);\\
\For{each $K \in KSet$}{   
     \If {\texttt{abs\_KNN\_predict\_same}($T, n, K, x, y$) = $False$} {
	\Return unknown; 
     }
}
\Return certified;\\
}
\end{algorithm}

In the \emph{abstract} learning step (Line~3), instead of considering $T$, our method considers the set of all clean datasets in $\DBIAS{}_n(T)$ symbolically, to compute the set of possible optimal $K$ values, denoted $KSet$.

In the \emph{abstract} prediction step (Lines~4-8), for each $K$, instead of considering input $x$, our method considers all perturbed inputs in $\Delta^\epsilon(x)$ and all clean datasets in $\DBIAS{}_n(T)$ symbolically, to check if the classification output always stays the same.  Our method returns ``certified'' only when  the classification output always stays the same (Line~9);  otherwise, it returns ``unknown'' (Line~6).

\textcolor{black}{
We only perturb numerical attributes in the input $x$ since perturbing categorical or binary attributes often does not make sense in practice.
}

In the next two sections, we present our detailed algorithms for abstracting the prediction step and the learning step, respectively.

\section{Abstracting the KNN Prediction Step}
\label{sec:abs-knn-prediction}

We start with abstract KNN prediction, which is captured by the subroutine \textit{\texttt{abs\_KNN\_predict\_same}} used in Line~5 of Algorithm~\ref{alg:overview}.  It consists of two parts. 
The first part (to be presented in Section \ref{subsec:abs-nn}) computes a superset of $T_x^K$, denoted $overNN$, while considering the impact of $\epsilon$ perturbation of the input $x$.  
The second part (to be presented in Section \ref{subsec:abs-label}) leverages $overNN$ to decide if the classification output always stays the same, while considering the impact of label-flipping bias in the dataset $T$.

\subsection{Finding the $K$-Nearest Neighbors}
\label{subsec:abs-nn}

To compute $overNN$, which is a set of samples in $T$ that \emph{may be} the $K$ nearest neighbors of the test input $x$, we must be able to compute the distance between $x$ and each sample in $T$.

This is not a problem at all in the concrete domain, since the $K$ nearest neighbors of $x$ in $T$, denoted $T_x^K$, is fixed and is determined solely by the Euclidean distance between $x$ and each sample in $T$ in the attribute space.  
However, when $\epsilon$ perturbation is applied to $x$, the distance changes and, as a result, the $K$ nearest neighbors of $x$ may also change.

Fortunately, the distance in the attribute space is not affected by label-flipping bias in the dataset $T$, since label-flipping only impacts sample labels, not sample attributes.
Thus, in this subsection, we only need to consider the impact of $\epsilon$ perturbation of the input $x$.

\subsubsection{The Challenge.}

Due to $\epsilon$ perturbation,  a single test input $x$ becomes a potentially-infinite set of inputs $\Delta^\epsilon(x)$. 
Since our goal is to over-approximate the $K$ nearest neighbors of $\Delta^\epsilon(x)$, the expectation is that, as long as there exists some $x' \in \Delta^\epsilon(x)$ such that a sample input $t$ in $T$ is one of the $K$ nearest neighbors of $x'$, denoted $t\in T_{x'}^K$, we must include $t$ in the set $overNN$.  
That is,
\[
   \bigcup_{x' \in \Delta^\epsilon(x)} T_{x'}^K \subseteq overNN \subseteq T .
\] 
However, finding an efficient way of computing $overNN$ is a challenging task.  
As explained before, the naive approach of enumerating $x' \in \Delta^\epsilon(x)$, computing the $K$ nearest neighbors, $T_{x'}^K$, and unionizing all of them would not work.
Instead, we need abstraction that is both efficient and accurate enough in practice.

Our solution is that, for each sample $t$ in $T$, we first analyze the distances between $t$ and all inputs in  $\Delta^\epsilon(x)$ symbolically, to compute a lower bound and an upper bound of the distances.
Then, we leverage these lower and upper bounds to compute the set $overNN$, which is a superset of samples in $T$ that may become the $K$ nearest neighbors of $\Delta^\epsilon(x)$.

\subsubsection{Bounding Distance Between $\Delta^\epsilon(x)$ and $t$.}

Assume that $x = (x_1, x_2, ..., x_D)$ and $t = (t_1, t_2, ..., t_D)$ are two real-valued vectors in the $D$-dimensional attribute space.  
Let $\epsilon = (\epsilon_1, \epsilon_2, ..., \epsilon_D)$, where $\epsilon_i \geq 0$, be the small perturbation. 
Thus, the perturbed input is $x' = (x'_1, x'_2, ..., x'_D) = (x_1+\delta_1, x_2+\delta_2, ..., x_D+\delta_D)$, where  $\delta_i \in [-\epsilon_i,\epsilon_i]$ for all $i=1,...,D$.

The distance between $x$ and $t$ is a fixed value 
$d(x,t) = \sqrt{\sum_{i = 1}^D( x_i - t_i)^2}$,
since both $x$ and the samples $t$ in $T$ are fixed, 
but the distance between $x'\in \Delta^\epsilon(x)$ and $t$ is a function of $\delta_i \in [-\epsilon_i, \epsilon_i]$, since 
$\sqrt{\sum_{i = 1}^D(x'_i - t_i)^2} = \sqrt{\sum_{i = 1}^D(x_i - t_i + \delta_i)^2}$. 
For ease of presentation, we define the distance as  $d^\epsilon = \sqrt{ \sum_{i=1}^{D} d_i^\epsilon }$, where $d_i^\epsilon = (x_i-t_i+\delta_i)^2$ is the (squared) distance function in the $i$-th dimension.   
Then, our goal becomes computing the lower bound, $LB(d^\epsilon)$, and the upper bound, $UB(d^\epsilon)$, in the domain $\delta_i \in [-\epsilon_i, \epsilon_i]$ for all $i=1,...,D$.

\subsubsection{Distance Bounds are Compositional.}

Our first observation is that bounds on the distance $d^\epsilon$ as a whole can be computed using bounds in the individual dimensions. 
To see why this is the case, consider the (square) distance in the $i$-th dimension, $d_i^{\epsilon} = (x_i - t_i + \delta_i)^2$, where $\delta_i \in [-\epsilon_i, \epsilon_i]$, and the (square) distance in the $j$-th dimension, $d_j^\epsilon  = (x_j - t_j + \delta_j)^2$, where $\delta_j \in [-\epsilon_j, \epsilon_j]$. 
By definition, $d_i^{\epsilon}$ is completely independent of $d_j^{\epsilon}$ when $i \neq j$.

Thus, the lower bound of $d^{\epsilon}$, denoted $LB(d^\epsilon)$, can be calculated by finding the lower bound of each $d_i^{\epsilon}$ in the $i$-th dimension. 
Similarly, the upper bound of $d^{\epsilon}$, denoted $UB(d^\epsilon)$, can also be calculated by finding the upper bound of each $d_i^{\epsilon}$ in the $i$-the dimension.  That is,

$LB(d^\epsilon) = \sqrt{\sum_{i = 1}^D LB(d_i^\epsilon)}$ and 
$UB(d^\epsilon) = \sqrt{\sum_{i = 1}^D UB(d_i^\epsilon)}$.

\subsubsection{Four Cases in Each Dimension.}

Our second observation is that, by utilizing the mathematical nature of the (square) distance function, we can calculate the minimum and maximum values of $d_i^\epsilon$, which can then be used as the lower bound $LB(d_i^\epsilon)$ and upper bound $UB(d_i^\epsilon)$, respectively.

Specifically, in the $i$-th dimension, 
%
the (square) distance function $d_i^{\epsilon} = ((x_i-t_i) + \delta_i)^2$ may be rewritten to  $(\delta_i+A)^2$, where $A = (x_i-t_i)$ is a constant and $\delta_i\in[-\epsilon,+\epsilon]$ is a variable. 
The function can be plotted in two dimensional space, using $\delta_i$ as $x$-axis and the output of the function as $y$-axis;  thus, it is a quadratic function $Y = (X+A)^2$.

\begin{figure}[t]
\centering
\begin{minipage}{0.49\linewidth}
\centering
\includegraphics[width=.5\textwidth]{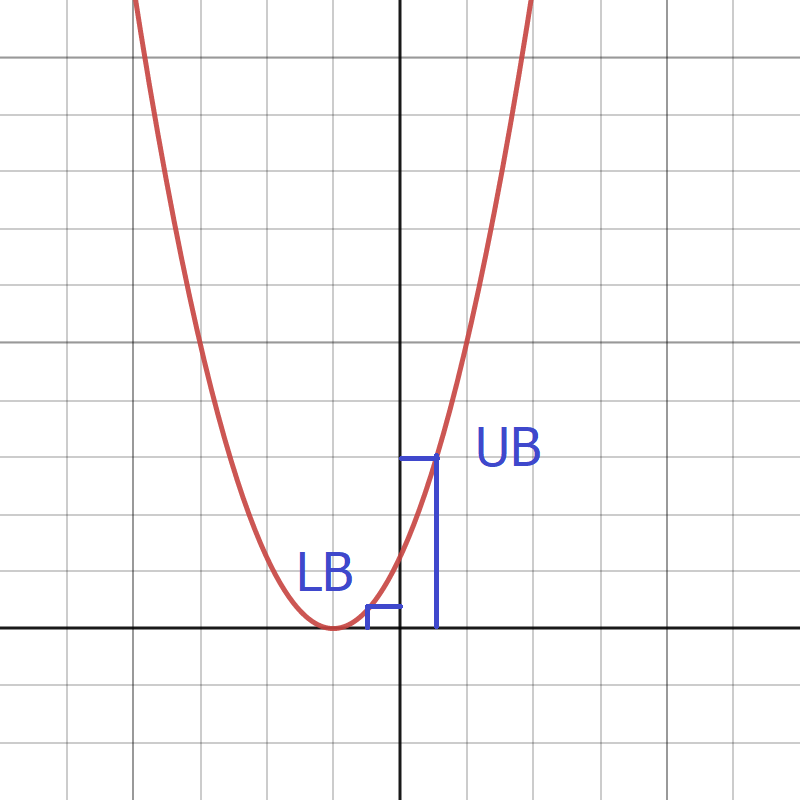}

(a)
\end{minipage}
\begin{minipage}{0.49\linewidth}
\centering
\includegraphics[width=.5\textwidth]{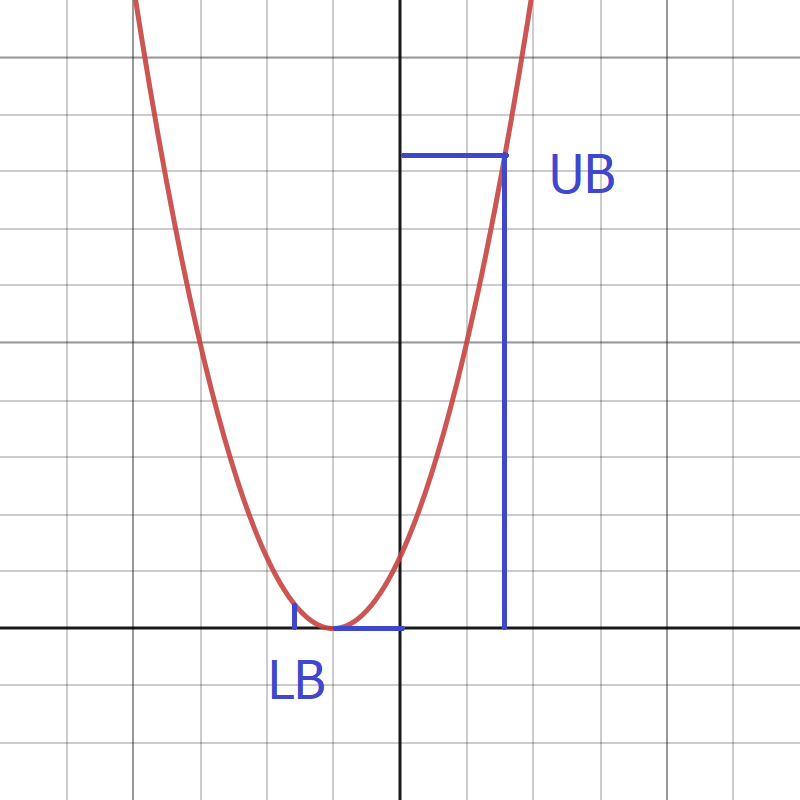}

(b)
\end{minipage}

\vspace{2ex}

\begin{minipage}{0.49\linewidth}
\centering
\includegraphics[width=.5\textwidth]{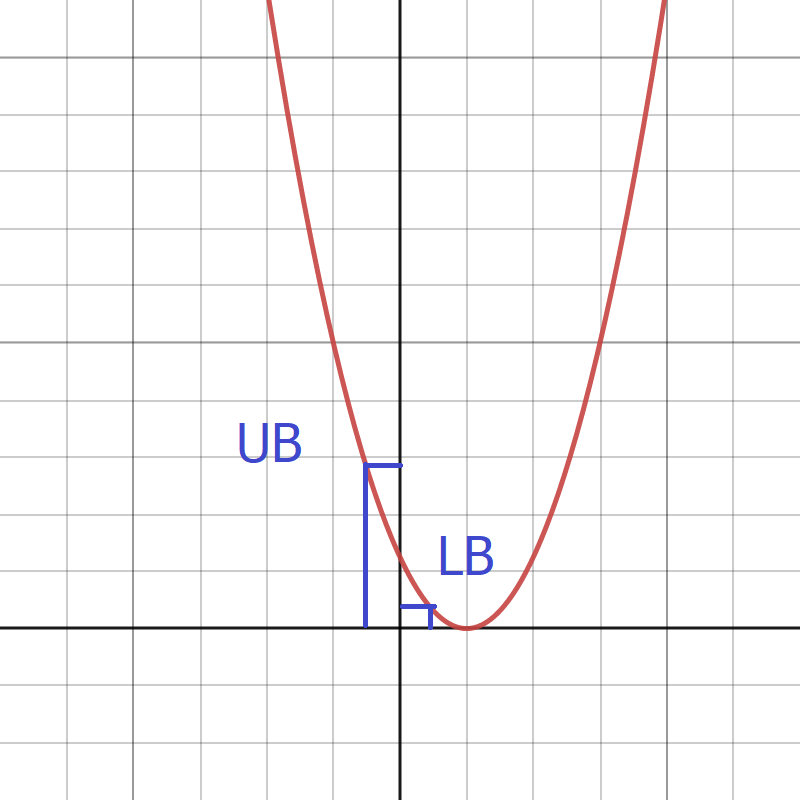}

(c)
\end{minipage}
\begin{minipage}{0.49\linewidth}
\centering
\includegraphics[width=.5\textwidth]{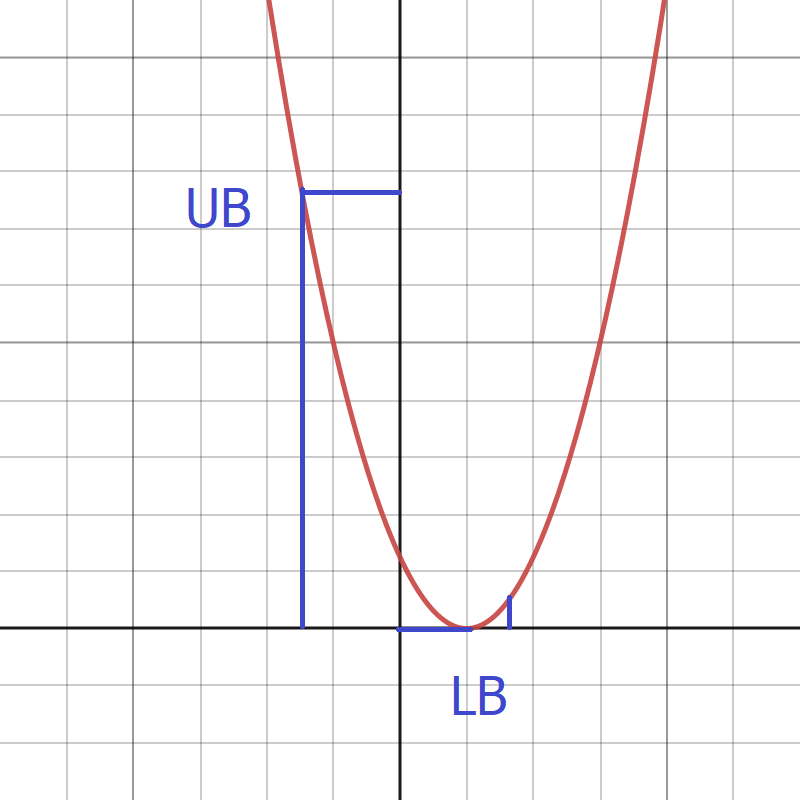}

(d)
\end{minipage}

\caption{Four cases for computing the upper and lower bounds of the distance function $d_i^\epsilon(\delta_i) = (\delta_i + A)^2$ for $\delta_i\in [-\epsilon_i,\epsilon_i]$. 
In these figures, $\delta_i$ is the $x$-axis, and $d_i^\epsilon$ is the $y$-axis, \textsf{LB} denotes $LB(d_i^\epsilon)$, and \textsf{UB} denotes $UB(d_i^\epsilon)$. }
\label{fig:four-cases}
\end{figure}

Fig.~\ref{fig:four-cases} shows the plot, which reminds us of where the minimum and maximum values of a quadratic function is. 
There are two versions of the quadratic function, depending on whether $A>0$ (corresponding to the two subfigures at the top) or $A<0$ (corresponding to the two subfigures at the bottom).  
Each version also has two cases, depending on whether the perturbation interval $[-\epsilon_{i},\epsilon_{i}]$ falls inside the constant interval $[-|A|, |A|]$ (corresponding to the two subfigures on the left)  or falls outside (corresponding to the two subfigures on the right). 
Thus, there are four cases in total.

In each case, the maximal and minimal values of the quadratic function are different, as shown by the LB and UB marks in Fig.~\ref{fig:four-cases}.

\paragraph{Case (a)}

This is when $(x_i-t_i)>0$ and $-\epsilon_{i} > - (x_i-t_i)$, which is the same as saying $A>0$ and $-\epsilon_{i} >-A$.  In this case, function $d_i(\epsilon_i) = (\delta_i+ A)^2$ is monotonically increasing w.r.t. variable $\delta_i\in [-\epsilon_{i}, +\epsilon_{i}]$. 

Thus, $LB(d_i^{\epsilon}) = (-\epsilon_{i} + (x_i-t_i))^2$ and 
$UB(d_i^{\epsilon}) = (+\epsilon_{i} + (x_i-t_i))^2$.

\paragraph{Case (b)}

This is when $(x_i-t_i)>0$ and $-\epsilon_{i} < - (x_i-t_i)$, which is the same as saying $A>0$ and $-\epsilon_{i} <-A$.  In this case, the function is not monotonic.
The minimal value is 0, obtained when $\delta_i=-A$. 
The maximal value is obtained when $\delta_i= +\epsilon_{i}$.

Thus, $LB(d_i^{\epsilon}) = 0$ and 
$UB(d_i^{\epsilon}) = (+\epsilon_{i} + (x_i-t_i))^2$.

\paragraph{Case (c)}

This is when $(x_i-t_i)<0$ and $\epsilon_{i} < - (x_i-t_i)$, which is the same as saying  $A<0$ and $\epsilon_{i} <-A$.  In this case, the function is monotonically decreasing w.r.t. variable $\delta_i\in [-\epsilon_{i}, \epsilon_{i}]$.

Thus, $LB(d_i^{\epsilon}) = (\epsilon_{i} + (x_i-t_i))^2 $ and 
$UB(d_i^{\epsilon}) = (-\epsilon_{i} + (x_i-t_i))^2$.

\paragraph{Case (d)}

This is when $(x_i-t_i)<0$ and $\epsilon_{i} > - (x_i-t_i)$, which is the same as saying $A<0$ and $\epsilon_{i} >-A$. In this case, the function is not monotonic. 
The minimal value is 0, obtained when $\delta_i=-A$. 
The maximal value is obtained when $\delta_i= - \epsilon_{i}$.

Thus, $LB(d_i^{\epsilon}) = 0$ and
$UB(d_i^{\epsilon}) = (-\epsilon_{i} + (x_i-t_i))^2$.

\paragraph{Summary}

By combining the above four cases, we compute the bounds of  the entire distance function $d^\epsilon$ as follows:
\[
 \left[~~ \sqrt{ \sum_{i = 1}^D\max(|x_i - t_i| - \epsilon_i, 0 )^2}, ~~~~ \sqrt{ \sum_{i = 1}^D(|x_i - t_i| + \epsilon_i)^2} ~~ \right]
\]
Here, the take-away message is that, since $x_i$, $t_i$ and $\epsilon_i$ are all fixed values,  the upper and lower bounds can be computed in constant time, despite that there is a potentially-infinite number of inputs in $\Delta^\epsilon (x)$.

\subsubsection{Computing $overNN$ Using Bounds.}

With the upper and lower bounds of the distance between $\Delta^\epsilon(x)$ and sample $t$ in the dataset $T$, denoted $[LB(d^\epsilon(x,t))$, $UB(d^\epsilon(x,t))]$, we are ready to compute  $overNN$ such that every $t\in overNN$ \emph{may be} among the $K$ nearest neighbors of $\Delta^\epsilon(x)$.

Let $UB_{Kmin}$ denote the $K$-th minimum value of $UB(d^\epsilon(x,t))$ for all $t \in T$. 
Then, we define $overNN$ as the set of samples in $T$ whose $LB(d^\epsilon(x,t))$ is \emph{not greater than} $UB_{Kmin}$. In other words, 
\[
overNN = \{ t \in T  ~|~ LB(d^\epsilon(x,t)) \leq UB_{Kmin} \}.
 \]

\paragraph{Example}  
Given a dataset $T= \{t^1, t^2, t^3, t^4, t^5\}$, a test input $x$, perturbation $\epsilon$, and $K=3$. 
Assume that the lower and upper bounds of the distance between $\Delta^\epsilon(x)$ and samples in $T$  are $[25.4, 29.4]$, $[30.1, 34.1]$, $[35.3, 39.3]$, $[37.2, 41.2]$, $[85.5, 90.5]$. 
Since $K=3$, we find the 3rd minimum upper bound,  $UB_{3min} = 39.3$. 
By comparing $UB_{3min}$ with the lower bounds, we compute $overNN_3 = \{t^1, t^2, t^3, t^4 \}$,
since $t^5$ is the only sample in $T$ whose lower bound is greater than $39.3$. 
All the other four samples \emph{may be} among the 3 nearest neighbors of $\Delta^\epsilon(x)$.

Due to $\epsilon$ perturbation, the set $overNN_3$ for $K=3$ is expected to contain 3 or more samples.  That is, since different inputs in $\Delta^\epsilon(x)$ may have different samples as their 3-nearest neighbors, to be conservative, we have to take the union of all possible sets of 3-nearest neighbors.

\paragraph{Soundness Proof} 
Here we prove that any $t' \notin overNN_K$ cannot be among the K nearest neighbors of any $x' \in \Delta^\epsilon(x)$.
Since $UB_{Kmin}$ is the $K$-th minimum $UB(d^\epsilon(x,t))$ for all $t \in T$, there must be samples $t^1, t^2, ... t^K$ such that $UB(d^\epsilon(x,t^i)) \leq UB_{Kmin}$ for all $i = 1, 2, ...K$. 
For any $t' \notin overNN$, we have $LB(d^\epsilon(x,t')) > UB_{Kmin}$. 

Combining the above conditions, we have $LB(d^\epsilon(x,t')) > UB(d^\epsilon(x,t^i))$ for $i = 1, 2, ... K$. It means at least $K$ other samples are closer to $x$ than $t'$. 
Thus, $t'$ cannot be among the $K$-nearest neighbors of $x'$.

\subsection{Checking the Classification Result}
\label{subsec:abs-label}

Next, we try to certify that, regardless of which of the $K$ elements are selected from $overNN$, the prediction result obtained using them is always the same. 

The prediction label is affected by both $\epsilon$ perturbation of  the input $x$ and label-flipping bias in the dataset $T$.  Since $\epsilon$ perturbation affects which points are identified as the $K$ nearest neighbors, and its impact has been accounted for by $overNN$, from now on, we focus only on label-flipping bias in $T$.

\begin{algorithm}[t]
\caption{Subroutine $\texttt{abs\_same\_label}(overNN, K, y)$.}
\label{alg:abs-label}
Let $S$ be a subset of $overNN$ obtained by removing all $y$-labeled elements;\\
Let $y' = Freq(S)$, and $\#y'$ be the count of $y'$-labeled elements in $S$;\\
\If {$\#y' <  K - |S| - 2 * n$} {
	\Return $True$;
}
\Return  $False$;
\end{algorithm}

Our method is shown in Algorithm \ref{alg:abs-label}, which takes the set $overNN$, the parameter $K$, and the expected label $y$ as input, and checks if it is possible to find a subset of $overNN$ with size $K$, whose most frequent label differs from $y$.  If such a ``bad'' subset cannot be found, we say that KNN prediction always returns the same label.

To try to find such a ``bad'' subset of $overNN$, we first remove all elements labeled with $y$ from $overNN$, to obtain the set $S$ (Line 1).  After that, there are two cases to consider. 
\begin{enumerate}
\item 
If the size of $S$ is equal to or greater than $K$, then any subset  of $S$ with size $K$ must have a different label because it will not contain any element labeled with $y$. 
Thus, the condition in Line 3 of Algorithm~\ref{alg:abs-label} is not satisfied (\#$y'$ is a positive number, and right-hand side is a negative number), and the procedure returns $False$.
\item 
If the size of $S$, denoted $|S|$, is smaller than $K$,  the most likely ``bad'' subset will be $S_K = S \cup \{$ any $(K - |S|)$ $y$-labeled elements from $overNN\}$. 
In this case, we need to check if the most frequent label in $S_K$ is $y$ or not. 
\end{enumerate}

In $S_K$, the most frequent label must be either $y$ (whose count is $K - |S|$) or $y'$ (which is the most frequent label in $S$, with the count $\#y'$).
Moreover, since we can flip up to $n$ labels, we can flip $n$ elements from label $y$ to label $y'$.

Therefore, to check if our method should return $True$, meaning the prediction result is guaranteed to be the same as label $y$,  we only need to compare $K - |S|$ with $\#y' + 2 * n$. 
This is checked using the condition in Line 3 of Algorithm~\ref{alg:abs-label}.

\section{Abstracting the KNN Learning Step}
\label{sec:abs-knn-learning}

In this section, we present our method for abstracting the learning step, which computes the optimal $K$ value based on $T$ and the impact of flipping at most $n$ labels. 
The output is a super set of possible optimal $K$ values, denoted $KSet$.

Algorithm \ref{alg:abs-knn-learn} shows our method, which takes the training set $T$ and parameter $n$ as input, and returns $KSet$ as output.
To be sound, we require the $KSet$ to include any candidate $k$ value that may become the optimal $K$ for some clean set $T'\in\DBIAS{}_n(T)$.

\begin{algorithm}[t]
\caption{Subroutine $abs\_KNN\_learn(T, n)$}
\label{alg:abs-knn-learn}
{\footnotesize
\For {each candidate $k$ value} { 
        Let $\{G_i\}$ = a partition of $T$ into $p$ groups of roughly equal size;\\
	$errUB_i^k$ = $\{ (x,y)\in G_i ~|~  \texttt{abs\_may\_err\ } 
                                                    (T\setminus G_i,n,k,x,y)=true \}$ for each $G_i$;\\
	$errLB_i^k$ = $\{ (x,y)\in G_i ~|~  \texttt{abs\_must\_err} 
                                                    (T\setminus G_i,n,k,x,y)=true \}$ for each $G_i$;\\
	$UB_k = \frac{1}{p}\sum_{i = 1}^{p} |errUB_i^k| / |G_i|$;\\
	$LB_k = \frac{1}{p}\sum_{i = 1}^{p} |errLB_i^k| / |G_i|$;\\
}
Let $minUB$ = $\min( \{UB_1, ..., UB_p\} )$;\\
\Return $KSet = \{ k ~|~ LB_k \leq minUB \}$;
}
\end{algorithm}

In Algorithm~\ref{alg:abs-knn-learn}, our method first computes the lower and upper bounds of the classification error for each $k$ value, denoted $LB_k$ and $UB_k$, as shown in Lines 5-6. 
Next, it computes $minUB$, which is the minimal upper bound for all candidate $k$ values (Line~8).
Finally, by comparing $minUB$ with $LB_k$ for each candidate $k$ value, our method decides whether this candidate $k$ value should be put into $KSet$ (Line~9).  

We will explain the steps needed to compute $LB_k$ and $UB_k$ in the remainder of this section.  
For now, assuming that they are available,  we explain how they are used to compute $KSet$.

\paragraph{Example}  
Given the candidate $k$ values,  $k_1, k_2, k_3, k_4$, and their error bounds  $[0.1, 0.2]$, $[0.1, 0.3]$, $[0.3, 0.4]$, $[0.3, 0.5]$. 
The smallest upper bound is $minUB = 0.2$.  By comparing $minUB$ with the lower bounds, we compute $KSet= \{k_1, k_2\}$, since only $LB_{k_1}$ and $LB_{k_2}$ are lower than or equal to $minUB$.

\paragraph{Soundness Proof}
Here we prove that any $k' \notin KSet$ cannot result in the smallest classification error.
Assume that $k_s$ is the candidate $k$ value that has the minimal upper bound ($minUB$), and $err_{k_s}$ is the actual classification error.  By definition, we have $err_{k_s} \leq minUB$. 
Meanwhile, for any $k' \notin KSet$, we have $LB_{k'} > minUB$.
Combining the two cases, we have $err_{k'} > minUB \geq err_{k_s}$. Here, $err_{k'} > err_{k_s}$ means that $k'$ cannot result in the smallest classification error.

\subsection{Overapproximating the Classification Error}

To compute the upper bound $errUB_i^k$ defined in Line~3 of Algorithm~\ref{alg:abs-knn-learn}, we use the subroutine \texttt{abs\_may\_err} to check if $(x,y)\in G_i$ may be misclassified when using $T\setminus G_i$ as the training set. 

Algorithm~\ref{alg:flip-UB-V2} shows the implementation of the subroutine, which checks, for a sample $(x,y)$, whether it is possible to obtain a set $S$ by flipping at most $n$ labels in $T_x^K$ such that the most frequent label in $S$ is not $y$. If it is possible to obtain such a set $S$, we conclude that the prediction label for $x$ may be an error. 

\begin{algorithm}[t]
\caption{Subroutine $\texttt{abs\_may\_err}(T, n, K, x, y)$.}
\label{alg:flip-UB-V2}
{\footnotesize
Let $y'$ be, among the  non-$y$ labels, the label with the highest count in $T_x^K$;\\
Let $\#y$ be the number of elements in $T_x^K$ with the $y$ label;\\
Let $n'$ be $\min(n, \#y \in T_x^K)$;\\
Changing $n'$ elements in $T_x^K$ from $y$ label to  $y'$ label;\\
\Return $Freq(T_x^K) \neq y$;\\
}
\end{algorithm}

The condition $Freq(T_x^K) \neq y$, computed on $T_x^K$ after the $y$ label of $n'$ elements is changed to $y'$ label, is a sufficient condition under which the prediction label for $x$ may be an error.
The rationale is as follows.

In order to make the most frequent label in the set $T_x^K$ different from $y$, we need to focus on the label most likely to become the new most frequent label.  It is the label $y' (\neq y)$ with the highest count in the current $T_x^K$.

Therefore, Algorithm \ref{alg:flip-UB-V2} checks whether $y'$ can become the most frequent label by changing at most $n$ elements in $T_x^K$ from $y$ label to  $y'$ label (Lines 3-5).

\subsection{Underapproximating the Classification Error}

To compute the lower  bound $errLB_i^k$ defined in Line~4 of Algorithm~\ref{alg:abs-knn-learn}, we use the subroutine \texttt{abs\_must\_err} to check if $(x,y)\in G_i$ must be misclassified when using $T\setminus G_i$ as the training set. 

Algorithm~\ref{alg:flip-LB} shows the implementation of the subroutine, which checks, for a sample $(x,y)$, whether it is impossible to obtain a set $S$ by flipping at most $n$ labels in $T_x^K$ such that the most frequent label in $S$ is $y$. 
In other words, is it impossible to avoid the classification error?   
If it is impossible to avoid the classification error, we conclude that the prediction label must be an error, and thus the procedure returns $True$

In this sense, all samples in $errLB_i^k$ (computed in Line~4 of Algorithm~\ref{alg:abs-knn-learn} are guaranteed to be misclassified.

\begin{algorithm}[t]
\caption{Subroutine $\texttt{abs\_must\_err}(T, n, K, x, y)$.}
\label{alg:flip-LB}
\If {$\exists S$ obtained from $T_x^K$ by flipping up to $n$ labels such that $Freq(S) = y$} {
	\Return $False$;
}
\Return $True$;
\end{algorithm}

The challenge in Algorithm \ref{alg:flip-LB} is to check if such a set $S$ can be constructed from $T_x^K$. 
The intuition is that, to make  $y$ the most frequent label, we should flip the labels of non-$y$ elements to label $y$. Let us consider two examples first.

\paragraph{Example 1}  
Given the label counts of $T_x^K$, denoted  \{$l_1$ * 4, $l_4$ * 4, $l_3$ * 2\}, meaning that 4 elements are labeled $l_1$, 4 elements are labeled $l_4$, and 2 elements are labeled $l_3$.  
Assume that  $n = 2$ and $y = l_3$. 
Since we can flip at most 2 elements, we choose to flip one $l_1 \rightarrow l_3$ and one $l_4 \rightarrow l_3$, to get a set $S$ = \{$l_1$ * 3, $l_4$ * 3, $l_3$ * 4\}.

\paragraph{Example 2}
Given the label counts of $T_x^K$, denoted  \{$l_1$ * 5, $l_4$ * 3, $l_3$ * 2\}, $n = 2$, and $y = l_3$.  We  can flip two $l_1 \rightarrow l_3$ to get a set $S$ = \{$l_1$ * 3, $l_4$ * 3, $l_3$ * 4\}.

\subsubsection{The LP Problem}

The question is how to decide whether the set $S$ (defined in Line~1 of Algorithm~\ref{alg:flip-LB}) exists.  
We can formulate it as a linear programming (LP) problem. 
The LP problem has two constraints.  The first one is defined as follows:
Let $y$ be the expected label, $l_i\neq y$ be another label, where $i=1,..., q$ and $q$ is the total number of class labels (e.g., in the above two examples, the number $q=3$). 
Let $\#y$ be the number of elements in $T_x^K$ that have the $y$ label.  Similarly, let $\#l_i$ be the number of elements with $l_i$ label.  
Assume that a set $S$ as defined in Algorithm~\ref{alg:flip-LB} exists, then all of the labels $l_i \neq y$  must satisfy 
\begin{equation}
\label{eq:LP}
\#l_{i} - \#flip_{{i}} < \#y + \sum_{i=1}^{q} \#flip_{i}  ~~,
\end{equation}
where $\#flip_{{i}}$ is a variable representing the number of  $l_{i}$--to--$y$ flips. 
Thus, in the above formula, the left-hand side is the count of $l_i$ after flipping, the right-hand side is the count of $y$ after flipping. Since $y$ is the most frequent label in $S$, $y$ should have a higher count than any other label. 

The second constraint is  
\begin{equation}
\sum_{i=1}^{q} \#flip_{i} \leq n  ~~,
\end{equation}
which says that the total number of label flips is bounded by the parameter $n$.

Since the number of class labels ($q$) is often small (from 2 to 10), this LP problem can be solved quickly.
However, the LP problem must be solved $|T|$ times, where $|T|$ may be as large as 50,000. 
To avoid invoking the LP solver unnecessarily, we propose two easy-to-check conditions. They are \emph{necessary} condition in that, if either of them is violated, the set $S$ does not exist.
Thus, we invoke the LP solver only if both conditions are satisfied.

\subsubsection{Necessary Conditions}

The first condition is derived from  Formula (1a), by adding up the two sides of the inequality constraint for all labels $l_i \neq y$.  The resulting condition is 
\[
\left(  \sum_{l_i\neq y} \#l_i - \sum_{i=1}^{q} \#flip_{i} \right) 
  < 
\left( (q - 1) \#y + (q - 1)\sum_{i=1}^{q} \#flip_{i}  \right)  .
\] 
The second condition requires that, in $S$, label $y$ has a higher count (after flipping) than any other label, including the label $l_p \neq y$ with the highest count in the current $T_x^K$. The resulting condition is 
\[
(\#l_p - \#y)/2 < n ,
\]
since only when this condition is satisfied, it is possible to allow $y$ to have a higher count  than $l_p$, by flipping at most $n$ of the label $l_p$ to $y$.

\textcolor{black}{
These are necessary conditions (but may not be sufficient conditions) because, whenever the first condition does not hold, Equation~(\ref{eq:LP}) does not hold either.  Similarly, whenever the second condition does not hold, Equation~(\ref{eq:LP}) does not hold either. In this sense, these two conditions are \emph{easy-to-check} over-approximations of Equation~(\ref{eq:LP}).
}

\section{Experiments}
\label{sec:experiment}

We have implemented our method as a software tool written in Python using the \textbf{scikit-learn} machine learning library.  
We evaluated our tool on six datasets that are widely used in the fairness research literature.

\subsubsection{Datasets}

Table~\ref{tbl:stats} shows the statistics of each dataset, including the name, a short description, the size ($|T|$), the number of attributes, the protected attributes, and the parameters $\epsilon$ and $n$.
The value of $\epsilon$ is set to 1\% of the attribute range.  
The bias parameter $n$ is set to 1 for small datasets, 10 for medium datasets, and 50 for large datasets.
The protected attributes include \emph{Gender} for all six datasets, and \emph{Race} for two datasets, \emph{Compas} and \emph{Adult}, which are consistent with known biases in these datasets.

\begin{table}[t]
\centering
\caption{Statistics of all of the datasets used during our experimental evaluation.}
\label{tbl:stats}
\scalebox{.75}{
\begin{tabular}{|l|l|c|c|c|c|}
\hline

Dataset\ \ \ \ \ &  Description & \ \ Size $|T|$\ \  	& \ \ \# Attr.\ \ 	&  \ Protected Attr.\    & Parameters $\epsilon$ and $n$  \\\hline
\hline
Salary 	& salary level~\cite{weisberg1985applied}     &  52    		& 4        	& Gender 				& $\epsilon= 1\%$ attribute range, $n=1$ \\\hline
Student & academic performance~\cite{cortez2008using}\ \ \  &  649   	& 30  	& Gender 				& $\epsilon= 1\%$ attribute range, $n=1$ \\\hline
German 	& credit risk~\cite{Dua:2019}                 &  1,000    	& 20     	& Gender			& $\epsilon= 1\%$ attribute range, $n=10$ \\\hline
Compas 	& recidivism risk~\cite{dieterich2016compas}  &  10,500  	& 16      	& \ \ Race+Gender\ \ 	& $\epsilon= 1\%$ attribute range, $n=10$ \\\hline
Default & loan default risk~\cite{yeh2009comparisons}	&  30,000    	& 36      	& Gender		& $\epsilon= 1\%$ attribute range, $n=50$ \\\hline
Adult	& earning power~\cite{Dua:2019}	& 48,842		& 14		& Race+Gender		& \ \ $\epsilon= 1\%$ attribute range, $n=50$\ \  \\\hline
\end{tabular}
}
\end{table}

In preparation for the experimental evaluation, we have employed state-of-the-art techniques in the machine learning literature to preprocess and balance the datasets for KNN, including encoding, standard scaling, k-bins-discretizer, downsampling and upweighting.

\subsubsection{Methods}

For comparison purposes, we implemented six variants of our method, by enabling or disabling the ability to certify label-flipping fairness, the ability to certify individual fairness, and the ability to certify $\epsilon$-fairness.  

Except for $\epsilon$-fairness, we also implemented the naive approach of enumerating all $T'\in\DBIAS{}_n(T)$.  Since the naive approach does not rely on approximation, its result can be regarded as the ground truth (i.e., whether the classification output for an input $x$ is truly fair).  Our goal is to obtain the ground truth on small datasets, and use it to evaluate the accuracy of our abstract interpretation based method. 
However, as explained before, enumeration does not work for  $\epsilon$-fairness, since the number of inputs in $\Delta^\epsilon(x)$ is infinite.

Our experiments were conducted on a computer with 2 GHz Quad-Core Intel Core i5 CPU and 16 GB of memory. 
The experiments were designed to answer two questions.  First, is our method efficient and accurate enough in handling popular datasets in the fairness literature?   
Second, does our method help us gain insights?  For example, it would be interesting to know whether decision made on an individuals from a protected minority group is more (or less) likely to be certified as fair.

\subsubsection{Results on Efficiency and Accuracy}

We first evaluate the efficiency and accuracy of our method. 
For the two small datasets, \emph{Salary} and \emph{Student}, we are able to obtain the ground truth using the naive enumeration approach, and then compare it with the result of our abstract interpretation based method. 
We want to know how much our results deviate from the ground truth.

\begin{table}[t]
\centering
\caption{Results for certifying \emph{label-flipping} and \emph{individual fairness} (gender) on small datasets, for which ground truth can still be obtained by naive enumeration, and compared with our method.}
\label{tbl:baseline-all}
\scalebox{.75}{
\begin{tabular}{|l|c|c|c|c|c|c|c|c|c|c|c|c|}
\hline
        &\multicolumn{6}{c|}{ Certifying label-flipping fairness }   
        &\multicolumn{6}{c|}{ \ \ \ Certifying label-flipping + individual fairness\ \ \  }   \\
\cline{2-13}
	& \ Ground\   &     & Our         &       &      	&  
        & \ Ground\   &     & Our         &       &     	&  \\
Name	& truth  &  Time    & \ method\   & Time  & \ Accuracy\	& \ Speedup\  
        & truth  &  Time    & \ method\   & Time  & \ Accuracy\ & \ Speedup\ 
\\
\hline\hline
Salary	
        & 50.0\%	& 1.7s		& 33.3\%	& 0.2s		&\textbf{66.7\%}	& 8.5X	
        & 33.3\%	& 1.5s		& 33.3\%	& 0.2s		&\textbf{100\%}		& 7.5X	
\\\hline
Student\ \
 	& 70.8\%	& 23.0s		& 60.0\%	& 0.2s		&\textbf{84.7\%}	& 115X
        & 58.5\%	& 25.2s		& 44.6\%	& 0.2s		&\textbf{76.2\%}	& 116X
\\\hline
\end{tabular}
}
\end{table}

Table~\ref{tbl:baseline-all} shows the results obtained by treating \emph{Gender} as the protected attribute. 
Column~1 shows the name of the dataset. 
Columns 2-7 compare the naive approach (ground truth) and our method in certifying label-flipping fairness.  
Columns 8-13 compare the naive approach (ground truth) and our method in certifying label-flipping plus individual fairness.

Based on the results in Table~\ref{tbl:baseline-all}, we conclude that the accuracy of our method is high (81.9\% on average) despite its aggressive use of abstraction to reduce the computational cost.
Our method is also 7.5X to 126X faster than the naive approach. 
Furthermore, the larger the dataset, the higher the speedup.

\textcolor{black}{
For medium and large datasets, it is infeasible for the naive enumeration approach to compute and show the ground truth in Table~\ref{tbl:baseline-all}. However, the fairness scores of our method shown in Table~\ref{tbl:label-all} provide ``lower bounds'' for the ground truth since our method is sound for certification.  For example, when our method reports 95\% for \emph{Compas (race)} in Table~\ref{tbl:label-all}, it means the ground truth must be $\geq$95\% (and thus the gap must be $\leq$5\%). However, there does not seem to be obvious relationship between the gap and the dataset size -- the gap may be due to some unique characterristics of each dataset.
}

\subsubsection{Results on the Certification Rates}

We now present the success rates of our certification method for the three variants of fairness.
Table~\ref{tbl:label-all} shows the results for label-flipping fairness in Columns~2-3, label-flipping plus individual fairness (denoted \emph{+ Individual fairness}) in Columns~4-5, and label-flipping plus $\epsilon$-fairness (denoted \emph{+ $\epsilon$-fairness}) in Columns~6-7.
For each variant of fairness, we show the percentage of test inputs that are certified to be fair, together with the average certification time (per test input).
In all six datasets, \emph{Gender} was treated as the protected attribute.  In addition, \emph{Race} was treated as the protected attribute for \emph{Compas} and \emph{Adult}.

\begin{table}[t]
\centering
\caption{Results for certifying \emph{label-flipping}, \emph{individual}, and \emph{$\epsilon$-fairness} by our method.}
\label{tbl:label-all}
\scalebox{.75}{
\begin{tabular}{|l|c|c|c|c|c|c|}
\hline
Name	         &  \ Label-flipping fairness\ \            & \ \ Time\ \  
                 &  \ + Individual fairness\ \          & \ \ Time\ \  
                 &  \ \ \ \ \ \ + $\epsilon$-fairness\ \ \ \ \ \  & \ \ Time\ \	\\\hline
\hline
Salary (gender)	     & 33.3\% & 0.2s  & 33.3\%  & 0.2s	& 33.3\%  & 0.2s	\\\hline
Student (gender)\ \  & 60.0\% & 0.2s  & 44.6\%  & 0.2s 	& 32.3\%  & 0.2s	\\\hline
German (gender)	     & 48.0\% & 0.2s  & 44.0\%  & 0.3s	& 43.0\%  & 0.2s	\\\hline
Compas (race)	     & 95.0\% & 0.3s  & 62.4\%  & 1.4s  & 56.4\%  & 1.1s	\\\hline
Compas (gender)	     & 95.0\% & 0.3s  & 65.3\%  & 1.3s	& 59.4\%  & 1.0s	\\\hline
Default (gender)     & 83.2\% & 2.3s  & 73.3\%  & 4.4s	& 64.4\%  & 3.5s	\\\hline
Adult (race)	     & 76.2\% & 2.2s  & 65.3\%  & 4.5s	& 53.5\%  & 5.3s	\\\hline
Adult (gender)	     & 76.2\% & 2.2s  & 52.5\%  & 3.5s 	& 43.6\%  & 3.3s	\\\hline
\end{tabular}
}
\end{table}

From the results in Table~\ref{tbl:label-all}, we see that as more stringent fairness standard is used,  the certified percentage either stays the same (as in \emph{Salary}) or decreases (as in \emph{Student}).   
This is consistent with what we expect, since the classification output is required to stay the same for an increasingly larger number of scenarios. 
For \emph{Compas (race)}, in particular, adding $\epsilon$-fairness on top of label-flipping fairness causes the certified percentage to drop from 62.4\% to 56.4\%.

Nevertheless, our method still maintains a high certification percentage.   Recall that, for \emph{Salary}, the 33.3\% certification rate (for \emph{+Individual fairness}) is actually 100\% accurate according to comparison with the ground truth in Table~\ref{tbl:baseline-all}, while the 44.6\% certification rate (for \emph{+Individual fairness}) is actually 76.2\% accurate.
Furthermore, the efficiency of our method is high: for \emph{Adult}, which has 50,000 samples in the training set, the average certification time of our method remains within a few seconds.

\subsubsection{Results on Demographic Groups}

Table~\ref{tbl:group} shows the certified percentage of each demographic group, when both \emph{label-flipping} and \emph{$\epsilon$-fairness} are considered, and both \emph{Race} and \emph{Gender} are treated as protected attributes. 
The four demographic groups are (1) \emph{White Male}, (2) \emph{White Female}, (3) \emph{Other Male}, and (4) \emph{Other Female}. 
For each group, we show the certified percentage obtained by our method. 
In addition, we show the weighted averages for \emph{White} and \emph{Other}, as well as the weighted averages for \emph{Male} and \emph{Female}.

\begin{table}[t]
\centering
\caption{Results for certifying  \emph{label-flipping + $\epsilon$-fairness} with both \emph{Race} and \emph{Gender} as protected attributes.}
\label{tbl:group}
\scalebox{.75}{
{\footnotesize (a) Compas\ \ \ }
\begin{tabular}{|l|c|c|c|}
\hline
		& \ \ White\ \ 	& \ \ Other\ \ 		& Wt. Avg		\\\hline
\hline
Male	        & 61.9\%    	& 52.2\%		& 52.8\%	 \\\hline
Female  	& 100\%		& 60.0\%		& 63.7\%   \\\hline
Wt. Avg\ \	& 63.7\%	& 53.7\%		& 54.4\%	\\\hline
\end{tabular}
\quad
\hspace{5ex}
{\footnotesize (b) Adult\ \ \ }
\begin{tabular}{|l|c|c|c|}
\hline
		& \ \ White\ \ 	& \ \ Other\ \ 		& Wt. Avg		\\\hline
\hline
Male    	& 35.3\%    	& 33.3\%		& 35.1\%	 \\\hline
Female   	& 33.3\%	& 66.7\%		& 37.0\%   \\\hline
Wt. Avg\ \  	& 34.7\%	& 44.4\%		& 35.6\%	\\\hline
\end{tabular}
}
\end{table}

For \emph{Compas}, \emph{White Female} has the highest certified percentage (100\%) while \emph{Other Female} has the lowest certified percentage (52.2\%); here, the classification output represents the recidivism risk.

For \emph{Adult}, \emph{Other Female} has the highest certified percentage (66.7\%) while the other three groups have certified percentages in the range of 33.3\%-35.3\%.

The differences may be attributed to two sources, one of which is technical and the other is social. 
The social reason is related to historical bias, which is well documented for these datasets.  If the actual percentages (ground truth) is different, the percentages reported by our method will also be different. 
The technical reason is related to the nature of the KNN algorithm itself, which we explain as follows.

In these datasets, some demographic groups have significantly more samples than others.   In KNN, the lowest occurring group may have a limited number of close neighbors. Thus, for each test input $x$ from this group, its $K$ nearest neighbors tend to have a larger radius in the input vector space.  As a result, the impact of $\epsilon$ perturbation on $x$ will be smaller, resulting in fewer changes to its $K$ nearest neighbors.
That may be one of the reasons why, in Table~\ref{tbl:group}, the lowest occurring groups,  \emph{White Female} in \emph{Compas} and \emph{Other Female} in \emph{Adult}, have significantly higher certified percentage than other groups.

\textcolor{black}{
Results in Table~\ref{tbl:group} show that, even if a machine learning technique discriminates against certain demographic groups, for an individual, the prediction result produced by the machine learning technique may still be fair. 
This is closely related to differences (and sometimes conflicts) between \emph{group fairness} and \emph{individual fairness}: while group fairness focuses on statistical parity, individual fairness focuses on similar outcomes for similar individuals. Both are useful notions and in many cases they are complementary.
}

\subsubsection{Caveat}

Our work should not be construed as an endorsement nor criticism of the use of machine learning techniques in socially sensitive applications.
Instead, it should be viewed as an effort on developing new methods and tools to help improve our understanding of these techniques.

\section{Related Work}
\label{sec:related}

For fairness certification, as explained earlier in this paper, our method is the first method for certifying KNN in the presence of historical (dataset) bias.  While there are other KNN certification and falsification techniques, including Jia et al.~\cite{jia2022certified} and Li et al.~\cite{li2022proving,li2023falsify}, they focus solely on robustness against data poisoning attacks as opposed to individual and $\epsilon$-fairness against historical bias.  Meyer et al.~\cite{MeyerAD21,abs-2206-03575} and Drews~ et al.~\cite{Drews2020PLDI} propose certification techniques that handle dataset bias, but target different machine learning techniques (decision tree or linear regression); furthermore, they do not handle $\epsilon$-fairness.

\textcolor{black}{
Throughout this paper, we have assumed that the KNN learning (parameter-tuning) step is not tampered with or subjected to fairness violation. However, since the only impact of tampering with the KNN learning step will be changing the optimal value of the parameter $K$, the biased KNN learning step can be modeled using a properly over-approximated $KSet$.  With this new $KSet$, our method for certifying fairness of the prediction result (as presented in  Section~\ref{sec:abs-knn-prediction}) will work AS IS.
}


Our method aims to certify fairness with certainty. In contrast, there are statistical techniques that can be used to prove that a system is fair or robust with a high probability.
Such techniques have been applied to various machine learning models, for example, in \textit{VeriFair}~\cite{bastani2019probabilistic} and \textit{FairSquare}~\cite{albarghouthi2017fairsquare}.
However, they are typically applied to the prediction step while ignoring the learning step, although the learning step may be affected by dataset bias.


There are also techniques for mitigating bias in machine learning systems.  Some focus on improving the learning algorithms using random smoothing~\cite{RosenfeldWRK20}, better embedding~\cite{bolukbasi2016man} or fair representation~\cite{RuossBFV20}, while others rely on formal methods such as iterative constraint solving~\cite{wang2022synthesizing}.  There are also techniques for repairing models to improve fairness~\cite{albarghouthi2017repairing}. 
Except for Ruoss et al.~\cite{RuossBFV20}, most of them focus on group fairness such as demographic parity and equal opportunity; they are significantly different from our focus on certifying individual and $\epsilon$-fairness of the classification results in the presence of dataset bias.

\textcolor{black}{
At a high level, our method that leverages a sound over-approximate analysis to certify fairness can be viewed as an instance of the abstract interpretation paradigm~\cite{CousotC77}.  Abstract interpretation based techniques have been successfully used in many other settings, including verification of deep neural networks~\cite{GehrMDTCV18,PaulsenW22}, concurrent software~\cite{KusanoW17,Kusano016,SungKW17}, and cryptographic software~\cite{Wu019,WuGS018}.
}


Since fairness is a type of non-functional property, the verification/certification techniques are often significantly different from techniques used to verify/certify functional correctness. 
Instead, they are more closely related to techniques for verifying/certifying robustness~\cite{ChaudhuriGL12},  noninterference~\cite{BartheDR04}, and side-channel security~\cite{WangSRW21,WangSW19,GuoWW18,ZhangGSW18}, where a program is executed multiple times, each time for a different input drawn from a large (and sometimes infinite) set, to see if they all agree on the output.  At a high level, this is closely related to differential verification~\cite{MohammadinejadP21,PaulsenWW20,PaulsenWWW20}, synthesis of relational invariants~\cite{WangW22} and verification of hyper-properties~\cite{SousaD16,FinkbeinerHT19}.

\section{Conclusions}
\label{sec:conclusion}

We have presented a method for certifying the individual and $\epsilon$-fairness of the classification output of the KNN algorithm, under the assumption that the training dataset may have historical bias. 
Our method relies on abstract interpretation to soundly approximate the arithmetic computations in  the learning and prediction steps. 
Our experimental evaluation shows that the method is efficient in handling popular datasets from the fairness research literature and accurate enough in obtaining certifications for a large amount of test data. 
While this paper focuses on KNN only, as a future work, we plan to extend our method to other machine learning models.

\bibliographystyle{splncs04}
\bibliography{main}
\end{document}